%% file: main.tex
\documentclass[10pt,twocolumn,letterpaper]{article}

\usepackage{cvpr}              

\input{preamble}

\usepackage{indentfirst}

\usepackage{array}
\usepackage{graphicx}
\usepackage{tabularx}
\usepackage{amsmath}
\usepackage{amssymb}
\usepackage{booktabs}
\usepackage{multirow}
\usepackage{tikz}

\input{acronym}

\definecolor{cvprblue}{rgb}{0.21,0.49,0.74}
\usepackage[pagebackref,breaklinks,colorlinks,citecolor=cvprblue]{hyperref}

\def\paperID{10} 
\def\confName{EDGE'24}
\def\confYear{2024}

\title{\ours: On-Device Text-to-Image Generation}

\author{Thibault Castells\textsuperscript{\dag} \quad Hyoung-Kyu Song\textsuperscript{\dag} \quad Tairen Piao\textsuperscript{\dag} \quad Shinkook Choi\textsuperscript{\dag}\\ Bo-Kyeong Kim\textsuperscript{\dag} \quad Hanyoung Yim\textsuperscript{\ddag} \quad Changgwun Lee\textsuperscript{\ddag} \quad Jae Gon Kim\textsuperscript{\ddag} \quad Tae-Ho Kim\textsuperscript{\dag}\\
{\tt\small \{thibault, hyoungkyu.song, tairen.piao, shinkook.choi, bokyeong.kim, thkim\}@nota.ai}\\
{\tt\small \{hanyoung.yim, cgwun.lee, jgon.kim\}@samsung.com}\\
\textsuperscript{\dag}Nota Inc., Korea \quad \textsuperscript{\ddag}Samsung Electronics, Korea
}

\begin{document}
\maketitle

\input{sec/0_abstract}

\input{sec/1_intro}
\input{sec/2_method}
\input{sec/3_experiment}
\input{sec/4_results}

\input{sec/5_conclusion}
{
    \small
    \bibliographystyle{ieeenat_fullname}
    \bibliography{main}
}

\input{sec/X_suppl}

\end{document}

%% file: preamble.tex
\usepackage[dvipsnames,table,xcdraw]{xcolor}

\usepackage{algorithm}
\usepackage{algorithmic}
\usepackage{tabularray}
\usepackage{subcaption}
\usepackage[utf8]{inputenc}
\usepackage{booktabs}
\usepackage{tabularx}
\usepackage{threeparttable}
\usepackage{amsmath}
\usepackage{amssymb}
\usepackage{xcolor}
\usepackage{pifont}
\usepackage{multirow}
\usepackage{colortbl}

\usepackage{threeparttable}
\newcommand{\cmark}{\textcolor{blue}{\checkmark}}
\newcommand{\xmark}{\textcolor{black}{\ding{55}}}
\usepackage{adjustbox}

%% file: acronym.tex
\newcommand{\ours}{EdgeFusion\xspace}
\newcommand{\stablediffusion}{SD\xspace}
\newcommand{\bksdm}{BK-SDM\xspace}
\newcommand{\bklcm}{BK-LCM\xspace}
\newcommand{\bksdmtiny}{BK-SDM-Tiny\xspace}
\newcommand{\bksdmadvtiny}{BK-SDM-Adv-Tiny\xspace}
\newcommand{\realvis}{Realistic\_Vision\_v5.1\xspace}

%% file: sec/0_abstract.tex
\begin{abstract}

The intensive computational burden of Stable Diffusion (SD) for text-to-image generation poses a significant hurdle for its practical application. To tackle this challenge, recent research focuses on methods to reduce sampling steps, such as Latent Consistency Model (LCM), and on employing architectural optimizations, including pruning and knowledge distillation. Diverging from existing approaches, we uniquely start with a compact SD variant, \bksdm. We observe that directly applying LCM to \bksdm with commonly used crawled datasets yields unsatisfactory results. It leads us to develop two strategies: (1) leveraging high-quality image-text pairs from leading generative models and (2) designing an advanced distillation process tailored for LCM. Through our thorough exploration of quantization, profiling, and on-device deployment, we achieve rapid generation of photo-realistic, text-aligned images in just two steps, with latency under one second on resource-limited edge devices.

\end{abstract}

%% file: sec/1_intro.tex
\section{Introduction}
\label{sec:intro}

Stable Diffusion (\stablediffusion) \cite{rombach2022high_stablediffusion} models have emerged as powerful tools in text-to-image (T2I) synthesis, acclaimed for their ability to transform text into high-quality images. These powerful generative models have widespread applications across various domains, from creative arts to practical solutions. However, their real-world deployment is hindered by immense computational and memory requirements due to complex iterative processes and large parameter sizes, making them difficult to deploy on resource-constrained devices like Neural Processing Units (NPUs).

To tackle this, existing studies focus on three directions: (1) \textit{architectural reduction} \cite{choi2023squeezing, li2023snapfusion, gupta2024progressive_ssd1b, chen2024pixartdelta, sauer2023adversarial_sdturbo_add, zhao2023mobilediffusion} to reduce the model size, (2) \textit{few-step inference} \cite{meng2023distillation, luo2023latent_lcm,li2023snapfusion,salimans2022progressive} to accelerate the inference speed, and (3) \textit{ leveraging AI-generated data} \cite{gupta2024progressive_ssd1b, chen2024pixartalpha} to improve the training efficiency. In particular, employing high-quality image-text pairs from advanced generative models \cite{he2024ptqd, gal2022image, kang2023scaling} overcomes limitations of real-world datasets' scarcity and biases while improving models' ability to produce visually compelling text-aligned images. However, the holistic integration of these approaches remains underexplored, especially regarding the practical deployment of compact SD models (see Tab.~\ref{tab:comparison}).

\input{figure/generated}

Adapting SD models in memory-constrained environments (e.g., NPUs), necessitates a targeted and optimized approach. This involves addressing the unique computational limitations and memory constraints of NPUs through a comprehensive strategy that includes architectural modifications and specialized optimizations specifically designed for these operational constraints \cite{dhariwal2021diffusion, li2024snapfusion}.

We propose EdgeFusion, a method that advances the field by optimizing \stablediffusion for execution on NPUs. 
EdgeFusion enhances the model efficiency by employing Block-removed Knowledge-distilled SDM (\bksdm)~\cite{kim2023architectural}, a foundational effort towards lightweight SD, and significantly improve its generation performance with superior image-text pairs from synthetic datasets. Furthermore, we refine the step distillation process of Latent Consistency Model (LCM)~\cite{luo2023latent_lcm} through empirical practices, achieving high-quality few-step inference. For deployment, EdgeFusion adopts model-level tiling, quantization, and graph optimization to generate a 512×512 image under one second on Samsung Exynos NPU~\cite{SamsungExynos2400}.

%% file: figure/generated.tex
\begin{figure}[t]
  \centering
  \includegraphics[width=\linewidth]{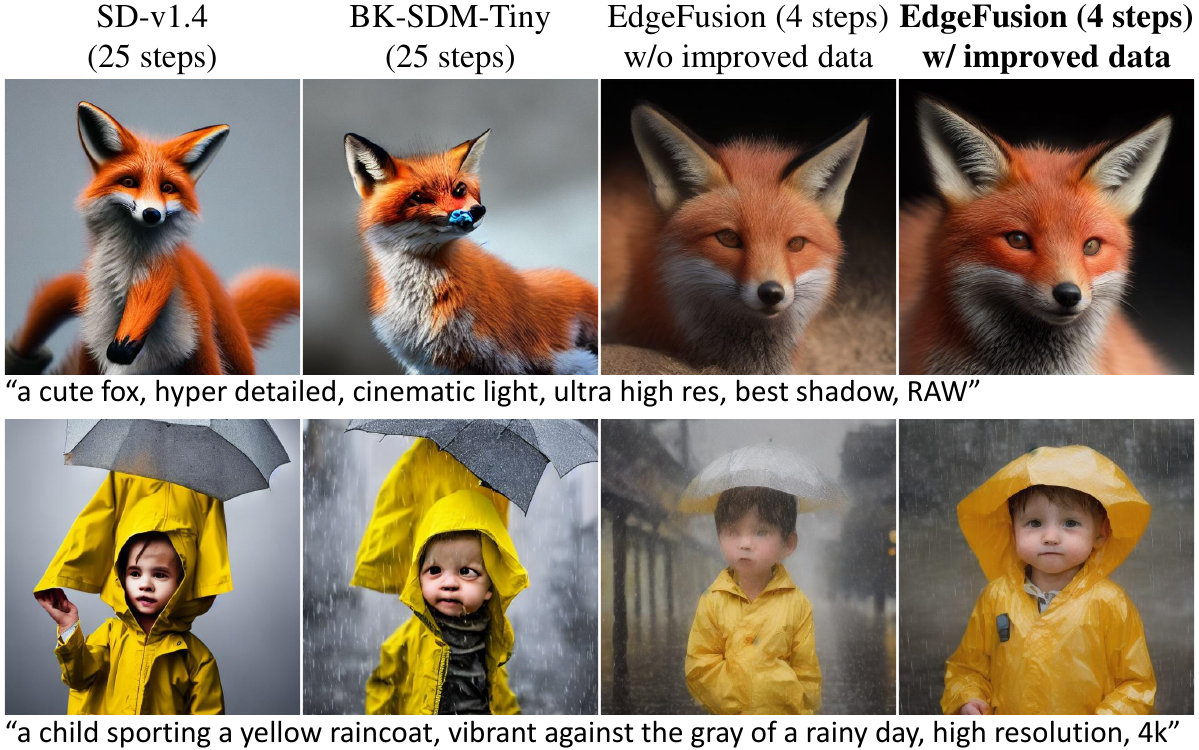}
    \vspace{-0.1in}
    \caption{\textbf{T2I generation results.} When trained with improved data, our EdgeFusion can produce high-quality images from challenging prompts in just a few denoising steps.}
    \label{fig:generated_main}
    \vspace{-0.1in}
\end{figure}

%% file: sec/2_method.tex
\input{table/comparison}

\section{Proposed approach}
\label{sec:method}

\subsection{Advanced distillation for LCM}
\label{sec:method-lcm}

\begin{figure}[t]
  \centering
  \includegraphics[width=\linewidth]{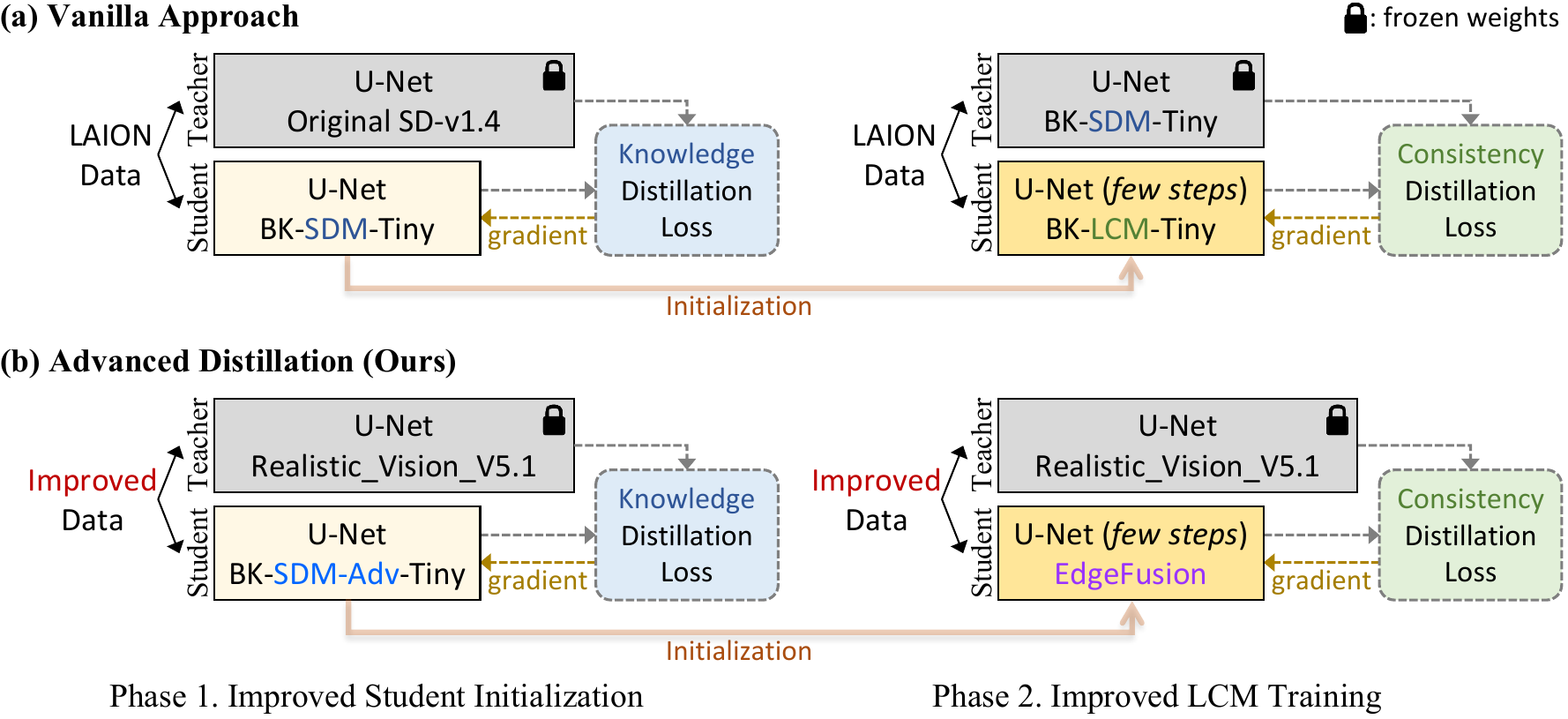}
       \vspace{-0.2in}
    \caption{\textbf{A compact SD with step reduction.} (a) Vanilla application of LCM: we initialize BK-LCM-Tiny with the weight from BK-SDM-Tiny and train with distillation to reduce sampling steps. (b) Our approach: improving the initialization of the LCM's student with a better teacher is beneficial. Moreover, in the LCM training phase, employing the original teacher enhances performance. Leveraging high-quality data is crucial in both phases.}\label{fig:method}
\end{figure}

To accelerate the generation speed, we collaboratively integrate multiple strategies. By adopting the compressed U-Net design of \bksdmtiny \cite{kim2023architectural}, we overcome the computational bottleneck inherent in the inference processes of diffusion-based T2I models. Additionally, by utilizing a recent step distillation method of LCM, plausible text-aligned images can be generated in as few as 2 or 4 denoising steps. However, we observed that the vanilla application of LCM's step reduction to the publicly accessible BK-SDM-Tiny checkpoint \cite{bk-sdm-tiny-url,kim2023architectural} (see Fig. \ref{fig:method}(a)) resulted in unsatisfactory outcomes.

To address this issue, we present our advanced distillation approach (see Fig. \ref{fig:method}(b)). In Kim~\etal~\cite{kim2023architectural}, the BK-SDM-Tiny was trained via feature-level knowledge distillation \cite{fitnets,heo2019comprehensive}, using the original SD-v1.4’s U-Net \cite{sdm_v1.4_hf,rombach2022high_stablediffusion} as its teacher. We empirically find that leveraging an advanced teacher model, namely Realistic\_Vision\_V5.1 \cite{realistic_vision_v5.1-url}, enhances the generation quality. This process leads to an improved student model, dubbed BK-SDM-Adv-Tiny.

Starting with BK-SDM-Adv-Tiny as the initialization, we fine-tune it with the scheduler in LCM to develop our fewer-step model, called EdgeFusion. Our experiments reveal notable improvements when utilizing the original large model as the teacher, instead of a model identically structured with the student. In both stages, the quality of the training data is crucial to achieving superior T2I diffusion models, which will be described further in the next section.

\subsection{Improved data quality}
\label{sec:method-data}

This section describes how we improve training data quality (dataset size and further details in Supplementary).

\vspace{+0.1cm}
\noindent \textbf{Initial training with LAION subset (L-orig).}
When the model was trained using the LAION data subset~\cite{laion_blog, laion5b} used in Kim~\etal~\cite{kim2023architectural}, the results were sub-optimal. We recognize a significant challenge with the LAION dataset: the prevalence of low-quality images and irrelevant captions. We hypothesize that such issues could greatly impede the training processes since capturing text-image relationships from noisy data is challenging and might demand considerably more extended training periods. It highlights the critical need to enhance the quality of the data.

\noindent \textbf{Data preprocessing (L-pp).}
To address the limitations of the LAION dataset, we perform data deduplication~\cite{penedo2023refinedweb}, remove samples smaller than 300 pixels, and utilize a model to optimize image cropping by selecting the best option among center crop and two crops from the image's longer side extremes, reducing the occurrence of cropped subjects. While these techniques yield better results, the gains in data quality are likely offset by the 44\% reduction in dataset size.

\noindent \textbf{Synthetic caption generation (L-pps).}
To enhance the caption informativeness, we use Sphinx~\cite{lin2023sphinx}, a large language model (LLM), to generate detailed, relevant captions for our preprocessed images, improving text-image correspondence. Caption generation is performed after image cropping to ensure prompts accurately reflect visual content. While synthetic captions have positive effects, they do not address image quality concerns from the initial data collection stage.

\noindent \textbf{Fully synthetic data generation (synt).}
To address persistent image quality issues in our dataset, we synthetically generate both images and corresponding prompts, providing higher control over training sample quality. Using GPT-4 \cite{openai2023gpt4} for prompt generation and SDXL \cite{podell2024sdxl} for image synthesis, we create a wide range of image-text pairs with notable improvements in image quality and text-image alignment.
This fully synthetic approach allows explicit control over generated image attributes like gender, ethnicity, age, and other demographic factors, creating a more diverse, inclusive dataset that mitigates potential biases and ensures comprehensive representation. We generate $62.3$k synthetic images using this approach. Including these high-quality, diverse, semantically aligned samples in our training set substantially improves our model's performance.

\noindent \textbf{Manual data curation (synt-cur).}
Despite significant improvements in image quality through fully synthetic data generation, some issues persist in the dataset, such as visual artifacts and suboptimal text-image alignment, particularly for complex or unusual prompts. To determine if the substantial human effort of manually curating synthetic data is justified, we curate $29.9$k synthetic images by correcting inexact prompts for improved image description accuracy and removing images containing artifacts.
\vspace{\baselineskip}

\subsection{Deployment on NPU}
\label{sec:method-deployment}

\noindent \textbf{Model-level tiling (MLT).}
Given the limited memory capacity of NPUs, dividing the model into smaller segments, called MLT, is crucial for efficient inference. FlashAttention~\cite{dao2022flashattention} introduces an algorithm capable of diminishing the data transfers between High Bandwidth Memory (HBM) and on-chip Static Random-Access Memory (SRAM), thereby enhancing performance within GPU environments. Motivated by this, we develop an MLT method to leverage heterogeneous computing resources, including tensor and vector engines, in edge devices. The primary objective of our tiling method is to reduce Dynamic Random-Access Memory (DRAM) access in handling Transformer blocks~\cite{dong2021attention} (e.g., due to large input dimensions of the Softmax layer). The secondary aim is to maximize SRAM utilization, as the smallest tile size could incur significant communication overhead~\cite{topcuoglu2002performance} between computing engines, necessitating an optimal tile count. We develop a method to determine the suitable number of tiles, based on calculations for the required size and utilization of SRAM.

\noindent \textbf{Quantization.}
To efficiently deploy \ours to Samsung Exynos NPU, we apply mixed-precision post-training quantization to the model. We apply FP16 precision for weights and activations to encoder and decoder modules, and apply INT8 weight quantization, INT16 activation quantization to the U-Net module. Sec.~\ref{sec:results-ablation} provides detailed quantization performance.

%% file: table/comparison.tex
\begin{table}[t]
\centering
\begin{adjustbox}{max width=\linewidth}

\begin{threeparttable}
\begin{tabular}{l|cccc|c}
\hline
Model                                            & \begin{tabular}[c]{@{}c@{}}Arch.\\Reduc.\textsuperscript{$\dagger$}\end{tabular} & \begin{tabular}[c]{@{}c@{}}AI-Gen.\\ Data\textsuperscript{$\dagger$}\end{tabular} & \begin{tabular}[c]{@{}c@{}}Few-Step\\Gen.\textsuperscript{$\dagger$} \end{tabular} & \begin{tabular}[c]{@{}c@{}}Edge\\ Deploy.\textsuperscript{$\dagger$}\end{tabular} & Param.\textsuperscript{$\dagger$}                        \\ \hline
Choi \etal \cite{choi2023squeezing}              & \cmark                                                          & \xmark                                                     & \xmark\,(20)                                                           & \cmark \textsuperscript{a}                                                 & -          \\
SnapFusion \cite{li2023snapfusion}               & \cmark                                                          & \xmark                                                     & \xmark\,(8)                                                      & \cmark \textsuperscript{b}                                                  & 1B         \\
LCM \cite{luo2023latent_lcm}                     & \xmark                                                           & \xmark                                                     & \cmark\,(2, 4)                                                 & \xmark                                                   & 1.3B+                           \\
SSD-1B \cite{gupta2024progressive_ssd1b}         & \cmark                                                          & \cmark                                                    & \xmark\,(25)                                 & \xmark                                                   & 1.3B+      \\
Pixart-alpha \cite{chen2024pixartalpha}          & \xmark                                                           & \cmark                                                    & \xmark\,(20)                                                     & \xmark                                                   & 0.6B        \\
Pixart-delta \cite{chen2024pixartdelta}          & \cmark                                                          & -                                                           & \cmark\,(1, 4)                                                   & \xmark                                                   & -                                \\
UFOGen \cite{xu2023ufogen}                       & \xmark                                                           & \xmark                                                     & \cmark\,(1)                                                      & \xmark                                                   & 0.9B        \\
SD-Turbo \cite{sauer2023adversarial_sdturbo_add} & \cmark                                                          & \xmark                     & \cmark\,(1)                                                      & \xmark                                                   & 0.86B \\
MobileDiff. \cite{zhao2023mobilediffusion}  & \cmark                                                          & \xmark                                                     & \cmark\,(1, 8)                                                      & \cmark \textsuperscript{c}                                                  & 0.4B                            \\
\rowcolor[HTML]{ECF4FF} 
\textbf{\ours (Ours)}                            & \cmark                                                        & \cmark                                                  & \cmark\,(1, 2, 4)    & \cmark \textsuperscript{d}   & 0.5B                             \\ \hline

\end{tabular}
\begin{tablenotes}[para,flushleft]
\normalsize
 \textsuperscript{$\dagger$}: Architectural Reduction; AI-Generated Data; Few-Step Generation (\#Steps); Edge Deployment; \#Parameters.
 \newline
\textsuperscript{a}: Galaxy S23. \textsuperscript{b}: iPhone14 Pro. \textsuperscript{c}: iPhone15 Pro.
\textsuperscript{d}: Samsung Exynos 2400.
\end{tablenotes}
\end{threeparttable}

\end{adjustbox}
\vspace*{-0.05in}
\caption{Comparison with existing work on efficient T2I synthesis.} \label{tab:comparison}

\end{table}

%% file: sec/3_experiment.tex
\section{Experimental setup}
\label{sec:experiment}

All training stages are conducted on a single NVIDIA A100 GPU. For \bksdm model training, a synthetic dataset is integrated alongside the existing training dataset while adhering to the established training recipe derived from prior studies. Likewise, the training process of applying LCM scheduler follows previous methods, ensuring fidelity to the established methodologies. 

For deployment on NPU devices, we fuse operators using a kernel fusion method~\cite{dao2022flashattention} and apply the MLT method. Experiments are conducted on Exynos 2400~\cite{SamsungExynos2400}, which features a 17K MAC NPU (2-GNPU + 2-SNPU) and DSP, to evaluate latency on edge devices.

Additionally, to assess the potential of the speed enhancements on cloud service, we provide a benchmark on a single NVIDIA A100 GPU.

%% file: sec/4_results.tex
\section{Results}
\label{sec:results}

\subsection{Main results}
\label{sec:results-main}

Tab.~\ref{tab:bksdm_eval} shows that despite being finetuned with just 29.9k samples (synt-cur), BK-SDM-Adv-Tiny outperforms \bksdmtiny by $5.28$ ($+17.4\%$) IS and $0.021$ CLIP scores. The lower FID score is likely due to the synthetic data distribution differing from COCO and \realvis having a worse FID than SD-v1.4. However, our human evaluation (Tab.~\ref{tab:human_eval}) showing a strong preference for BK-SDM-Adv-Tiny ($60.2\%$ win rate) suggests that while comparing to COCO is common practice, assuming COCO as the ideal data distribution may not align well with human preference.
Our final model, EdgeFusion, LCM-distilled from BK-SDM-Adv-Tiny on 182.2k samples (L-pps \& synt), has close performance to \bksdmtiny and even outperforms it on CLIP (Tabs.~\ref{tab:bksdm_eval}~and~\ref{tab:lcm_4step_eval}). Two steps results and more output examples can be found in the supplementary.

\input{table/eval_bksdm}
\subsection{Ablation studies}
\label{sec:results-ablation}

\noindent \textbf{Significance of high-quality training data.}
Tabs.~\ref{tab:bksdm_eval}~and~\ref{tab:lcm_4step_eval} show that synthetic prompts consistently improve text-image alignment (CLIP), but also affect the image quality: for LCM training, re-captioning LAION data improves both IS ($+1.99$) and FID ($-2.49$). Interestingly, the quality-quantity trade-off favors quality when finetuning BK-SDM-Tiny, with best results using synt-cur data only, while favoring quantity for LCM training, where combining L-pps and synt yields optimal results.
Tab.~\ref{tab:lcm_4step_eval} demonstrates the results of our advanced distillation pipeline (Fig.~\ref{fig:method}) with and without improved data, obtaining $+3.27$ IS, $-5.69$ FID, and $+0.018$ CLIP. Human evaluation in Tab.~\ref{tab:human_eval} confirms these observations with $62.8\%$ win rate for EdgeFusion.

\input{table/eval_4steps}

\noindent \textbf{Impact of fine-tuning the student first.} 
As shown in Tab.~\ref{tab:lcm_4step_eval}, removing \bksdmtiny finetuning from our advanced distillation pipeline results in a decrease in performance, with $-2.67$ IS, $+2.72$ FID and $-0.01$ CLIP. Again, this observation is confirmed by our human evaluation (Tab.~\ref{tab:human_eval}), with $61.7\%$ win rate for EdgeFusion.

\input{table/human_eval}

\noindent \textbf{Quantization.}
We compare the performance between FP32 and W8A16 quantized EdgeFusion. Fig. \ref{fig:quantization} shows the results of 3 prompts. There is no performance degradation after applying the W8A16 quantization.


\subsection{Model benchmark}
\label{sec:results-profiling}

\noindent \textbf{Inference speed on GPU} 
Tab.~\ref{tab:benchmark_gpu} compares the inference speed of EdgeFusion (without quantization) to BK-SDM-Tiny and SD-v1.4, demonstrating a remarkable $\times 10.3$ speedup compared to SD-v1.4.

\input{table/gpu_benchmark}

\noindent \textbf{NPU benchmark} 
Tab.~\ref{tab:relative_ratio_for_crossattention} shows the relative time ratio results for the cross attention block, without and with the MLT method. The results demonstrate a latency gain of approximately 73$\%$ when using MLT, effectively reducing Direct Memory Access (DMA) accesses.
Tab.~\ref{tab:benchmark} shows the inference efficiency of architectural elements in EdgeFusion.

\input{figure/quantization}

\input{table/relative_ratio_for_cross_attention}
\input{table/benchmark}

%% file: table/eval_bksdm.tex
\begin{table}[!h]
    \centering
    \footnotesize
    \begin{tabular}{@{}clccc@{}}
    \toprule
    KD teacher & Dataset & IS $\uparrow$ & FID $\downarrow$ & CLIP $\uparrow$ \\ 
    \arrayrulecolor{black} \midrule
    SD-v1.4~\cite{rombach2022high_stablediffusion} & L-orig \textsuperscript{a,~\cite{kim2023architectural}} & 30.39 & \textbf{17.43} & 0.266 \\
    \arrayrulecolor[gray]{0.7} \midrule \arrayrulecolor{black}
    \multirow{7}{*}{\begin{tabular}[c]{@{}c@{}}Realistic\\ Vision\\ v5.1~\cite{realistic_vision_v5.1-url} \end{tabular}} & L-orig \textsuperscript{b} & 32.11 & 21.66 & 0.278 \\
              & L-pp               & 32.20 & 20.98 & 0.278 \\
              & L-pps              & 32.32 & 21.65 & 0.283 \\
              & L-pps \& synt      & 34.82 & 23.37 & \textbf{0.288} \\
              & L-pps \& synt-cur  & 34.54 & 23.34 & \textbf{0.287} \\
              & synt               & 35.11 & 25.14 & \textbf{0.288} \\
              & \textbf{synt-cur} \textsuperscript{c}  & \textbf{35.67} & 26.15 & \textbf{0.287} \\
    \bottomrule
\end{tabular}

    Termed \textsuperscript{a}: w/o finetuning; \textsuperscript{b}: w/o improved data; \textsuperscript{c}: BK-SDM-Adv-Tiny.
    \caption{Results of the BK-SDM-Tiny architecture trained on different datasets. Evaluated on COCO dataset with $25$ steps.}
    \label{tab:bksdm_eval}
    \vspace{-0.15in}
\end{table}

%% file: table/eval_4steps.tex
\begin{table}[h]
    \centering
    \footnotesize
    
    \newcolumntype{C}{>{\centering\arraybackslash}X}
    \begin{tabular}{@{}llcccl@{}}
        \toprule
        Pretrained student & Dataset & IS $\uparrow$ & FID $\downarrow$ & CLIP $\uparrow$ \\ 
        \midrule
        BK-SDM-Tiny~\cite{kim2023architectural} & L-pps \& synt \textsuperscript{a} & 26.11 & 22.06 & 0.262 \\
        L-orig & L-orig \textsuperscript{b} & 25.51 & 25.03 & 0.254 \\
        L-pp & L-pp & 25.26 & 24.48 & 0.254 \\
        L-pps & L-pps & 27.25 & 21.99 & 0.266 \\
        L-pps \& synt & L-pps \& synt & 28.41 & \textbf{19.33} & \textbf{0.276} \\
        L-pps \& synt-cur & L-pps \& synt-cur & 28.11 & 20.07 & 0.273 \\
        synt & synt & 28.23 & 19.91 & 0.272 \\
        synt-cur & synt-cur & 27.87 & 20.21 & 0.270 \\
        \rowcolor[HTML]{ECF4FF} \textbf{synt-cur} & \textbf{L-pps \& synt} \textsuperscript{c} & \textbf{28.78} & \textbf{19.34} & 0.272 \\
        \bottomrule
    \end{tabular}

    Termed \textsuperscript{a}: w/o finetuning; \textsuperscript{b}: w/o improved data; \textsuperscript{c}: EdgeFusion.
    \vspace{-0.05in}
    \caption{Results for LCM-distillation of the BK-SDM-Tiny architecture from different pretrained students, with different datasets. Evaluated on COCO dataset with 4 inference steps.}

    \label{tab:lcm_4step_eval}
\end{table}


%% file: table/human_eval.tex
\begin{table}[h]
    \centering
    \footnotesize
    \begin{tabular}{lcc}
        \toprule
        Model & w/o finetuning & w/o improved data \\
        \midrule
        \text{\bksdmadvtiny} & 67.3\% & 60.9\% \\
        \text{\ours} & 61.7\% & 62.8\% \\
        \bottomrule
    \end{tabular}
      \vspace{-0.05in}
    \caption{Human preference evaluation. The win rate of our models against the same architecture without improved data and without student finetuning is reported (1500 comparisons, 21 participants).}
    \label{tab:human_eval}
      \vspace{-0.05in}
\end{table}

%% file: table/gpu_benchmark.tex
\begin{table}[h]
    \centering
    \footnotesize
    \begin{tabular}{lcr}
        \toprule
        Model & Inference Steps & Time (ms) \\
        \midrule
        SD-v1.4 & 25 & 930 \\
        \bksdmtiny & 25 & 540 \\
        \ours & 4 / 2 / 1 & 150 / 90 / 90 \\
        \bottomrule
    \end{tabular}
    \vspace{-0.05in}
    \caption{Benchmark results on NVIDIA A100 GPU (50 runs).}
    \label{tab:benchmark_gpu}

\end{table}

%% file: figure/quantization.tex
\begin{figure}[t]
    \centering
    \setlength{\tabcolsep}{2pt}
    \renewcommand{\arraystretch}{1.2}
    \newcommand{\imagewidth}{0.278\linewidth}
    \newcolumntype{x}{>{\centering\arraybackslash\hspace{0pt}}m{\imagewidth}}
    \newcolumntype{y}{>{\centering\arraybackslash\hspace{0pt}}m{0.1\linewidth}}
    \begin{tabular}{y x x x}
    {\scriptsize{FP32}} &
    \begin{tikzpicture}
    \node[anchor=south west,inner sep=0] at (0,0) {\includegraphics[trim=0px 0px {0px} 0px,clip,width=\linewidth]{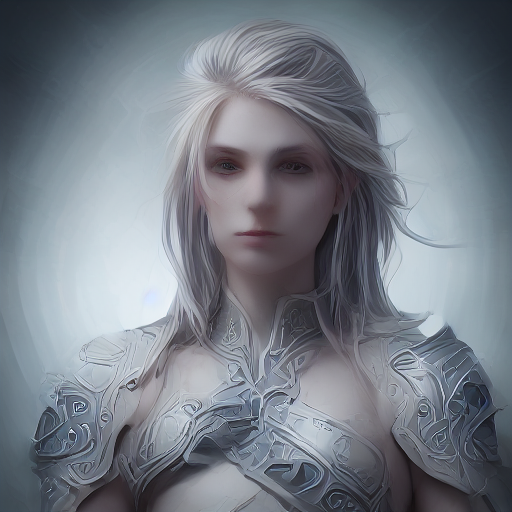}};
    \end{tikzpicture} &
    \begin{tikzpicture}
    \node[anchor=south west,inner sep=0] at (0,0) {\includegraphics[trim=0px 0px {0px} 0px,clip,width=\linewidth]{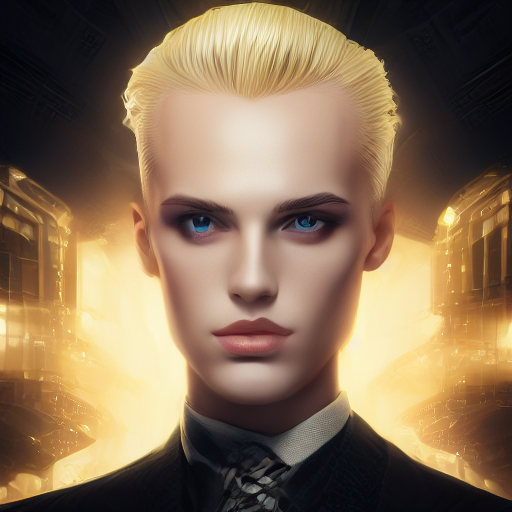}};
    \end{tikzpicture} &
    \begin{tikzpicture}
    \node[anchor=south west,inner sep=0] at (0,0) {\includegraphics[trim=0px 0px {0px} 0px,clip,width=\linewidth]{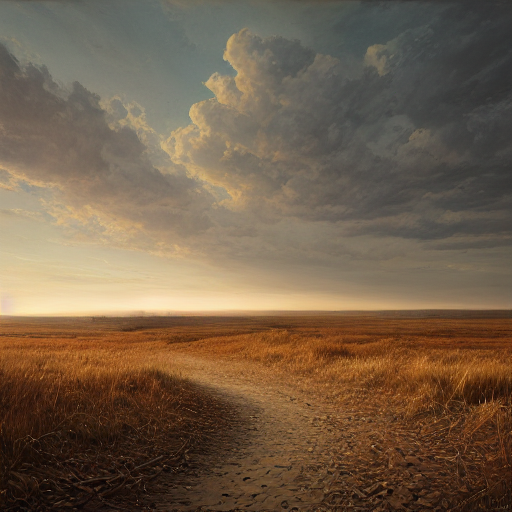}};
    \end{tikzpicture} \\[\dimexpr1pt]
    {\scriptsize{W8A16}} &
    \begin{tikzpicture}
    \node[anchor=south west,inner sep=0] at (0,0) {\includegraphics[trim=0px 0px {0px} 0px,clip,width=\linewidth]{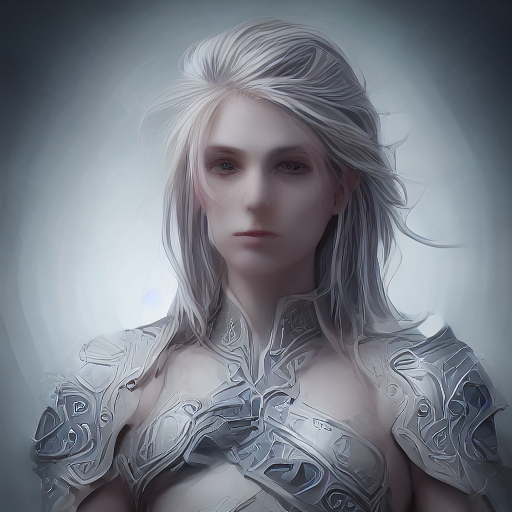}};
    \end{tikzpicture} &
    \begin{tikzpicture}
    \node[anchor=south west,inner sep=0] at (0,0) {\includegraphics[trim=0px 0px {0px} 0px,clip,width=\linewidth]{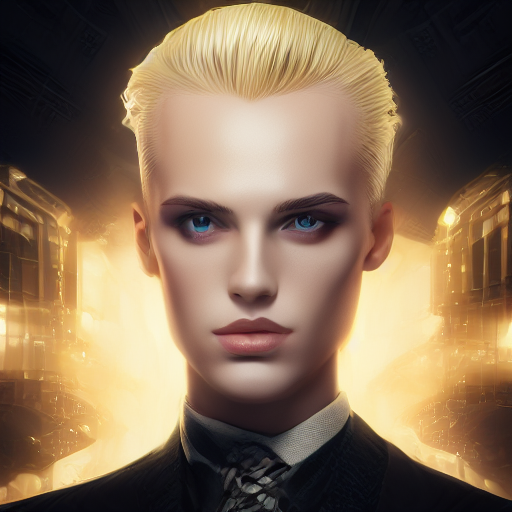}};
    \end{tikzpicture} &
    \begin{tikzpicture}
    \node[anchor=south west,inner sep=0] at (0,0) {\includegraphics[trim=0px 0px {0px} 0px,clip,width=\linewidth]{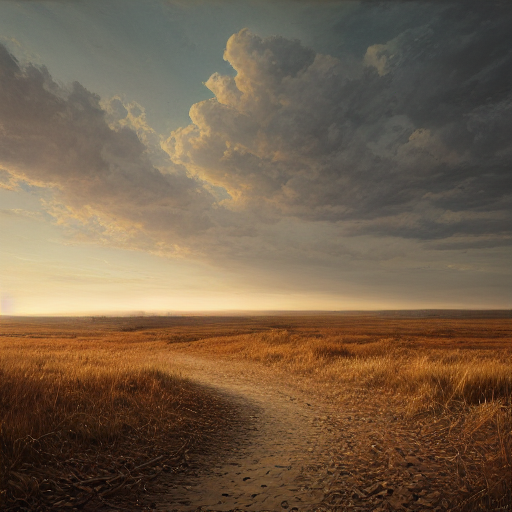}};
    \end{tikzpicture} \\
    \end{tabular}
    \vspace{-0.1in}
    \caption{\textbf{Comparison between FP32 and W8A16 quantized EdgeFusion (2 steps)}. Prompts can be found in Supplementary.}
    \label{fig:quantization}
\end{figure}

%% file: table/relative_ratio_for_cross_attention.tex
\begin{table}[!t]
    \centering
    \footnotesize
    
\begin{threeparttable}
    \begin{tabular}{lrr}
    \toprule
    Operation\textsuperscript{$\dagger$}                             & \multicolumn{1}{c}{w/o MLT} & \multicolumn{1}{c}{w/ MLT} \\ 
    \midrule
    DMA load for Query, Key               & 0.6\%                       & 0.6\%                      \\
    Tensor calculation: Query $\times$ Key$^T$ = S$_{in}$ & 6.9\%                    & 7.2\%                      \\
    DMA store: S$_{in}$                          & 19.1\%                      & -                          \\
    DMA load: S$_{in}$                           & 18.2\%                      & -                          \\
    Vector calculation: Softmax(S$_{in}$) = S$_{out}$      & 11.0\%                      & 11.6\%                     \\
    DMA store: S$_{out}$                          & 18.5\%                      & -                          \\
    DMA load: S$_{out}$, Value                    & 19.5\%                      & 0.8\%                      \\
    Tensor calculation: S$_{out}$ $\times$ Value = A        & 5.4\%                       & 5.7\%                      \\
    DMA store: A                          & 0.7\%                       & 0.7\%                      \\ \midrule
    Total                                 & 100\%                       & 27\%                       \\ 
    \bottomrule
\end{tabular}
\begin{tablenotes}[para,flushleft]
\footnotesize
 \textsuperscript{$\dagger$}: S$_{in}$ and S$_{out}$ indicate the input and output of the softmax layer, respectively. A is the output of the cross attention block.
\end{tablenotes}
\end{threeparttable}

\vspace{-0.05in}
    \caption{Inference time reduction using MLT. For the cross attention block, the relative time ratio is computed by comparing operations without MLT to those with MLT.
    } \label{tab:relative_ratio_for_crossattention}
\vspace{-0.05in}
\end{table}

%% file: table/benchmark.tex
\begin{table}[!t]
    \centering
    \footnotesize
    \centering
    \begin{tabular}{ccrrrr}
    \toprule
    \multirow{2}{*}{Model}    & \multirow{2}{*}{MLT}    & \multicolumn{4}{c}{Time (ms)}     \\
    \cmidrule(lr){3-6}
                              &                         & \multicolumn{1}{c}{Encoder} & \multicolumn{1}{c}{UNet}   & \multicolumn{1}{c}{Decoder} & \multicolumn{1}{c}{Total}  \\
    \midrule
    SDM v1.5                  & \xmark & 52.5    & 1350.3 & 483.4   & 1885.8 \\
    EdgeFusion                & \xmark & 51.9    & 454.3  & 484.7   & 990.9  \\
    \rowcolor[HTML]{ECF4FF} 
    EdgeFusion                & \cmark & 52.2    & 200.2  & 486.1   & 738.5  \\
    \bottomrule    
\end{tabular}

    \vspace{-0.05in}
    \caption{Benchmark on Exynos 2400 with and without MLT.}
    
    \label{tab:benchmark}
    \vspace{-0.1in}
\end{table}

%% file: sec/5_conclusion.tex
\section{Conclusion}
\label{sec:conclusion}


We introduce EdgeFusion to optimize SD models for efficient execution on resource-limited devices. Our findings show that utilizing informative synthetic image-text pairs is crucial for improving the performance of compact SD models. We also design an advanced distillation process for LCM to enable high-quality few-step inference. Through detailed exploration of deployment techniques, EdgeFusion can produce photorealistic images from prompts in one second on Samsung Exynos NPU.

%% file: sec/X_suppl.tex
\clearpage
\setcounter{page}{1}
\setcounter{section}{0}
\renewcommand*{\thesection}{\Alph{section}}
\maketitlesupplementary

\section{Data description}
\label{sec:data_description}

Tab.~\ref{tab:dataset_summary} summarize the datasets used in this paper, and precise the size of each dataset. Notably, our synthetic dataset (synt) is composed of $62.3k$ image-text pairs, and our manually curated synthetic dataset (synt-cur) is composed of $29.9k$ high quality image-text pairs.

\input{table/data_description}

\section{Teacher models}
\label{sec:teacher_models}

In Tab.~\ref{tab:eval_sdm}, we compare the performance of Realistic\_Vision\_5.1, used as a teacher model in this work, with SD-v1.4. While both models share the same architecture, their different training results in a gap in performance. Notably, Realistic\_Vision\_5.1 outperform SD-v1.4 on both IS and CLIP when evaluated on COCO dataset.

\input{table/eval_sdm}

\section{Supplementary results on data curation}
\label{sec:curated_vs_notcurated}

To better understand the effect of data curation, Tab.~\ref{tab:curated_vs_notcurated} provides a comparison of training using a subset of $10$k samples of our synthetic dataset (synt), and a manually curated version of this subset. The data curation removes $17.4\%$ of the data, resulting in $8.26$k samples in the curated dataset. The evaluation is performed for \bksdmtiny (25 steps), and \bklcm (2 and 4 steps), on COCO dataset.

\input{table/eval_curated_vs_notcurated}

\section{Supplementary output examples}
\label{sec:more_examples}

Fig.~\ref{fig:supp_bk_sdm_adv_tiny},~\ref{fig:supp_edgefusion_1step},~\ref{fig:supp_edgefusion_2step} and~\ref{fig:supp_edgefusion_4step} give additional output examples for \bksdmadvtiny and \ours with $1$, $2$ and $4$ diffusion steps, respectively. Prompts can be found in the figures descriptions.

\section{\ours 2-steps evaluation}
\label{sec:eval_2steps}

Tab.~\ref{tab:lcm_2step_eval} provides the evaluation on COCO dataset for BK-LCM-Tiny with 2 inference step.

\input{table/eval_2steps}

\input{figure/supp_bk_sdm_adv_tiny}

\input{figure/supp_edgefusion_1step}
\input{figure/supp_edgefusion_2step}

\input{figure/supp_edgefusion_4step}







\section{Prompts for FP32 and W8A16 Comparison}
\label{sec:prompts}

In this section, we show the prompts used in Fig.~\ref{fig:quantization}. Prompts: \textit{``cyclops fighter, white grey blue color palette, male, d\&d, fantasy, intricate, elegant, highly detailed, long silver hair, digital painting, artstation, octane render, concept art, matte, sharp focus, illustration, hearthstone, art by artgerm, alphonse mucha johannes voss''} (left), \textit{``Max Headroom in a Perfume advertisement, magical, science fiction, symmetrical face, large eyes, Chanel, Calvin Klein, Burberry, Versace, Gucci, Dior, hyper realistic, digital art, octane render, trending on artstation, artstationHD, artstationHQ, unreal engine, 4k, 8k''} (middle), \textit{``oil painting of holocaust LANDSCAPE, diffuse lighting, intricate, highly detailed, lifelike, photorealistic, illustration, concept art, smooth, sharp focus, art by francis bacon''} (right).

%% file: table/data_description.tex
\begin{table*}[!ht]
    \centering

    \begin{tabular}{l p{12.5cm} l}
        \toprule
        Dataset & Description & Size \\
        \midrule
        \midrule
        L-orig & LAION subset used to train BK-SDM-Tiny in~\citet{kim2023architectural} & 212.8k pairs \\
        \midrule
        L-pp & L-orig, plus additional preprocessing (deduplication, low resolution filtering, automatic cropping, manual removal) & 119.9k pairs \\
        \midrule
        L-pps & Recaptioned L-pp with synthetic prompts (generated using Sphinx~\cite{lin2023sphinx}) & 119.9k pairs \\
        \midrule
        synt & Synthetic text-image dataset, generated using GPT-4 and SDXL. Includes 21.6k manually curated pairs. & 62.3k pairs \\
        \midrule
        synt-cur & Manually curated synthetic text-image dataset, generated using GPT-4 and SDXL.  & 29.9k pairs \\
        \bottomrule
    \end{tabular}
        \vspace{-0.05in}
        \caption{Summary of datasets used in this study}
    \label{tab:dataset_summary}
        
\end{table*}

%% file: table/eval_sdm.tex
\begin{table}[!ht]
    \centering
    \footnotesize
    
    \newcolumntype{C}{>{\centering\arraybackslash}X}

    \begin{tabularx}{1\linewidth}{@{}lccc@{}}
        \toprule
        Teacher model & IS $\uparrow$ & FID $\downarrow$ & CLIP $\uparrow$ \\ 
        \midrule
        SD-v1.4 & 36.90 & \textbf{13.2} & 0.297 \\
        Realistic\_Vision\_v5.1 & \textbf{38.13} & 17.63 & \textbf{0.307} \\
        \bottomrule
    \end{tabularx}
        \vspace{-0.1in}
    \caption{Evaluation results for SD-v1.4 (used in BK-SDM~\cite{kim2023architectural}) and for Realistic\_Vision\_v5.1 (used in this work). The evaluation is performed on COCO dataset, with 25 inference steps.}
    \label{tab:eval_sdm}
        \vspace{-0.15in}
\end{table}

%% file: table/eval_curated_vs_notcurated.tex
\begin{table*}[!ht]
    \centering
    \footnotesize

    \begin{tabular}{l l c c c c}
        \toprule
        Model & Training dataset & Inference Steps & IS $\uparrow$ & FID $\downarrow$ & CLIP $\uparrow$ \\
        \midrule
        BK-SDM-Tiny & L-pps \& synt subset (not curated) & 25 & \textbf{33.37} & \textbf{22.58} & \textbf{0.285} \\
         & L-pps \& synt subset (curated) & 25 & 32.54 & 22.86 & \textbf{0.285} \\
        \midrule
        BK-LCM-Tiny & L-pps \& synt subset (not curated) & 4 & 27.52 & \textbf{21.33} & \textbf{0.270} \\
         & L-pps \& synt subset (curated) & 4 & \textbf{27.57} & 21.48 & 0.269 \\
        \midrule
        BK-LCM-Tiny & L-pps \& synt subset (not curated) & 2 & 26.31 & 22.52 & \textbf{0.265} \\
         & L-pps \& synt subset (curated) & 2 & \textbf{26.60} & \textbf{22.46} & \textbf{0.265} \\
        \bottomrule
    \end{tabular}
    \caption{IS, FID, and CLIP results for BK-SDM-Tiny and BK-LCM-Tiny models trained on L-pps and a subset of the synthetic image-pair dataset. The subset size is $10$k samples before being curated, and $8.26$k samples after.}
    \label{tab:curated_vs_notcurated}
    
\end{table*}

%% file: table/eval_2steps.tex
\begin{table}[ht]
    \centering
    \footnotesize
    
    \newcolumntype{C}{>{\centering\arraybackslash}X}
    \begin{tabular}{@{}llcccl@{}}
        \toprule
        Pretrained student & Dataset & IS $\uparrow$ & FID $\downarrow$ & CLIP $\uparrow$ \\ 
        \midrule
        BK-SDM-Tiny~\cite{kim2023architectural} & L-pps \& synt \textsuperscript{a} & 25.15 & 23.49 & 0.258 \\
        L-orig & L-orig \textsuperscript{b} & 23.72 & 27.30 & 0.248 \\
        L-pp & L-pp & 23.97 & 26.79 & 0.249 \\
        L-pps & L-pps & 25.93 & 23.36 & 0.261 \\
        L-pps \& synt & L-pps \& synt & 26.99 & 20.57 & \textbf{0.271} \\
        L-pps \& synt cur & L-pps \& synt cur & 26.63 & 21.20 & 0.268 \\
        synt & synth & 26.62 & 21.36 & 0.267 \\
        synt cur & synt cur & 26.53 & 21.41 & 0.264 \\
        \rowcolor[HTML]{ECF4FF} \textbf{synt cur} & \textbf{L-pps \& synt} \textsuperscript{c} & \textbf{27.11} & \textbf{20.47} & 0.268 \\
        \bottomrule
    \end{tabular}
    Termed \textsuperscript{a}: w/o finetuning; \textsuperscript{b}: w/o improved data; \textsuperscript{c}: EdgeFusion.
    \vspace{-0.05in}
    \caption{Results for LCM-distillation of the BK-SDM-Tiny architecture from different pretrained students, with different datasets. Evaluated on COCO dataset with 2 inference steps.}

    \label{tab:lcm_2step_eval}
\end{table}

%% file: figure/supp_bk_sdm_adv_tiny.tex
\begin{figure*}[t]
  \centering
  \includegraphics[width=\linewidth]{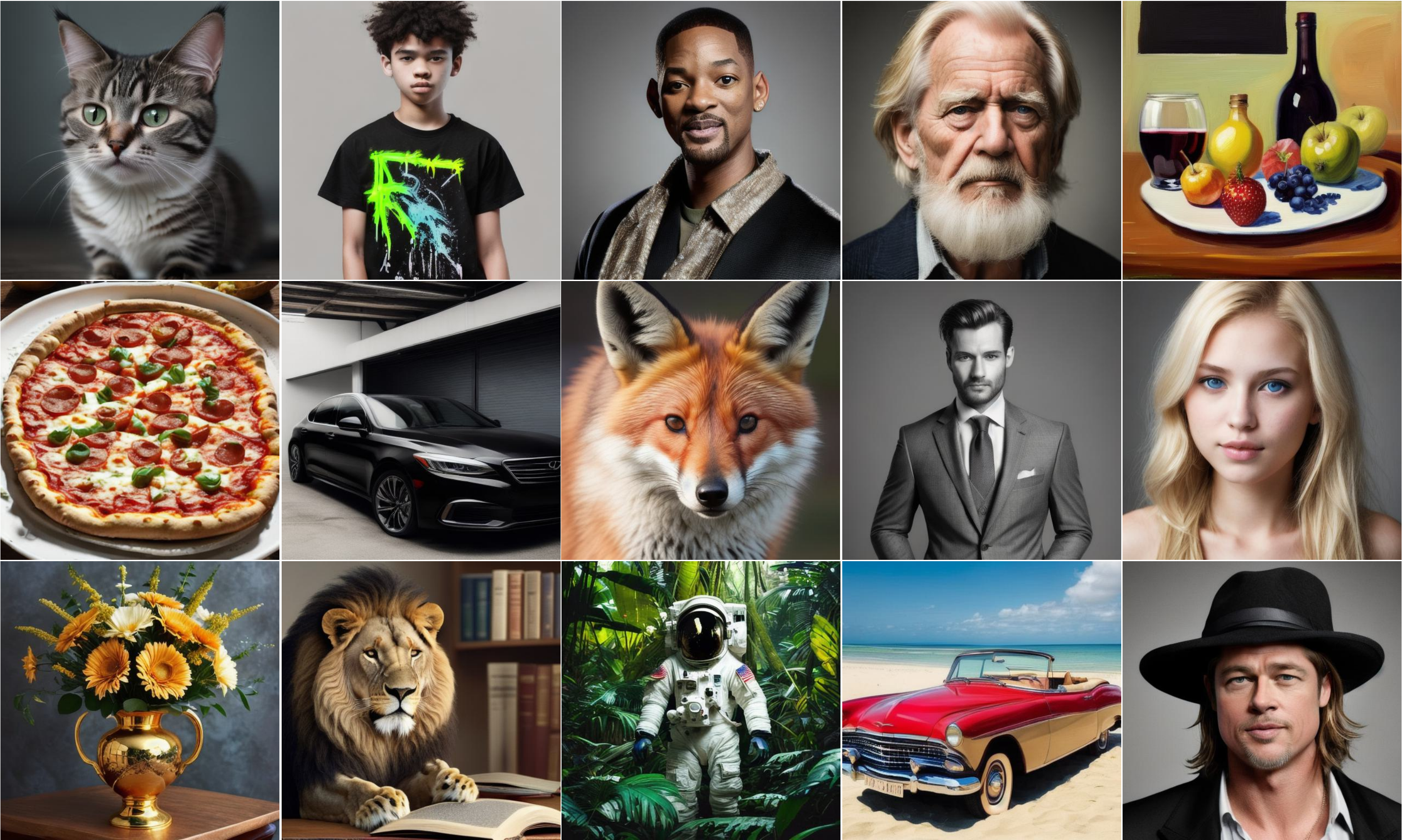}
       \vspace{-0.2in}
    \caption{\textbf{Output examples from \bksdmadvtiny.} Prompts (left to right, top to bottom): \textit{``a cute cat, hyper detailed, cinematic light, ultra high res, best shadow, RAW''}, \textit{``a teenager in a black graphic tee, the design a splash of neon colors''}, \textit{``photo portrait of Will Smith, hyper detailed, ultra high res, RAW''}, \textit{``photo portrait of an old man with blond hair''}, \textit{``painting of fruits on a table, next to a bottle of wine and a glass''}, \textit{``a beautiful pizza''}, \textit{``A sleek black sedan parked in a garage.''}, \textit{``a fox, photo-realistic, high-resolution, 4k''}, \textit{``photo portrait of a man wearing a charcoal gray suit, crisp and meticulously tailored''}, \textit{``photo portrait of a young woman with blond hair''}, \textit{``flowers in a golden vase''}, \textit{``a lion is reading a book on the desk, photo-realistic''}, \textit{``an astronaut in the jungle''}, \textit{``A classic convertible parked on a sunny beach.''}, \textit{``photo portrait of Brad Pitt wearing a black hat''}.}\label{fig:supp_bk_sdm_adv_tiny}
    \vspace{-0.05in}
\end{figure*}

%% file: figure/supp_edgefusion_1step.tex
\begin{figure*}[t]
  \centering
  \includegraphics[width=\linewidth]{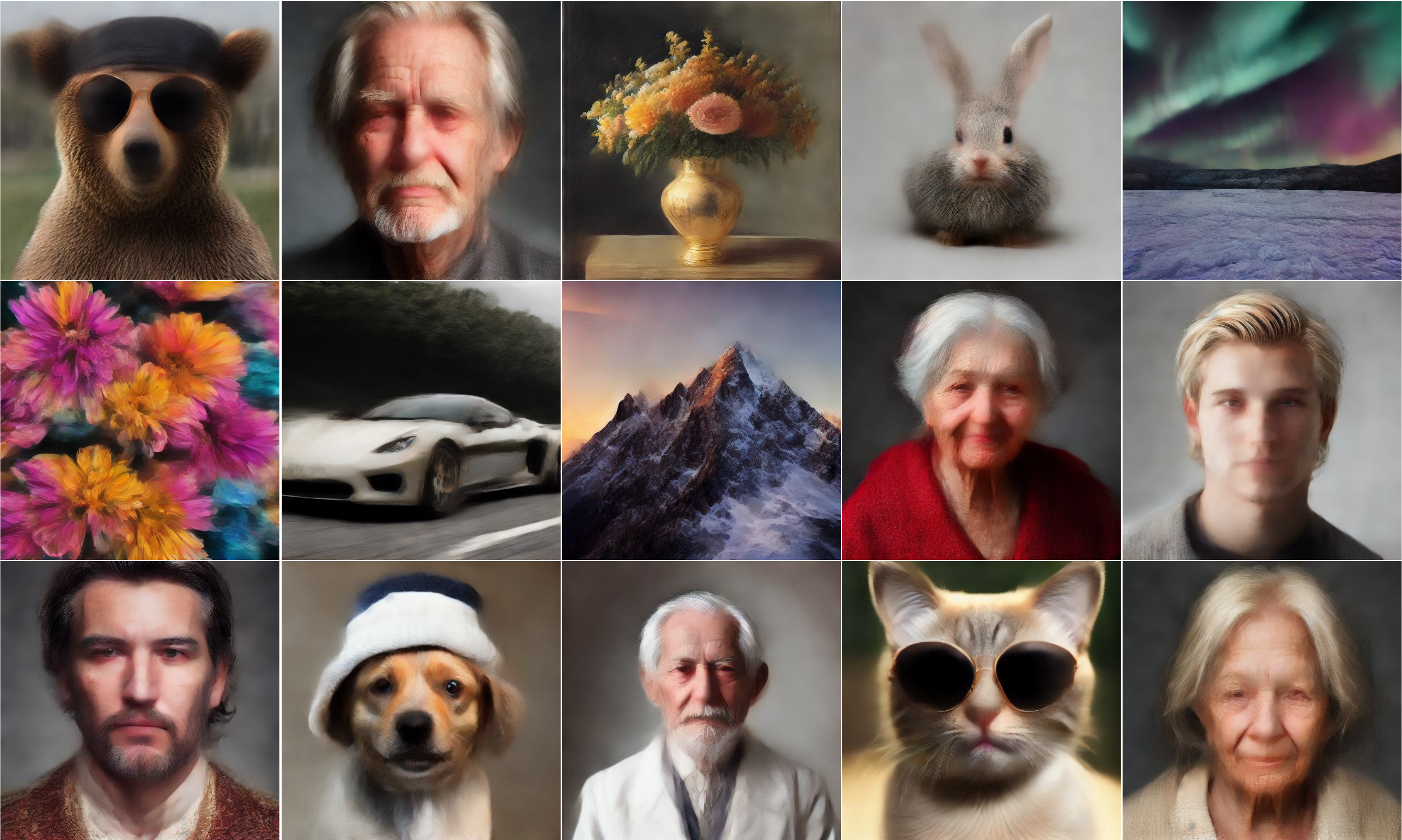}
       \vspace{-0.2in}
    \caption{\textbf{Output examples from \ours with 1 step inference.} Prompts (left to right, top to bottom): \textit{``a bear wearing a hat and sunglasses, photo-realistic, high-resolution, 4k''}, \textit{``photo portrait of an old man with blond hair''}, \textit{``flowers in a golden vase''}, \textit{``a cute rabbit''}, \textit{``a frozen tundra under the aurora borealis, the night sky alive with color, ice crystals sparkling like diamonds''}, \textit{``close up on colorful flowers''}, \textit{``a beautiful sports car''}, \textit{``a mountain peak during golden hour''}, \textit{``photo portrait of an old woman dressed in red''}, \textit{``photo portrait of a young man with blond hair''}, \textit{``beautiful photo portrait of the king, high resolution, 4k''}, \textit{``a cute dog wearing a hat''}, \textit{``photo portrait of an old man dressed in white''}, \textit{``a cute cat wearing sunglasses, photo-realistic''}, \textit{``photo portrait of an old woman with blond hair''}.}\label{fig:supp_edgefusion_1step}
\end{figure*}

%% file: figure/supp_edgefusion_2step.tex
\begin{figure*}[t]
  \centering
  \includegraphics[width=\linewidth]{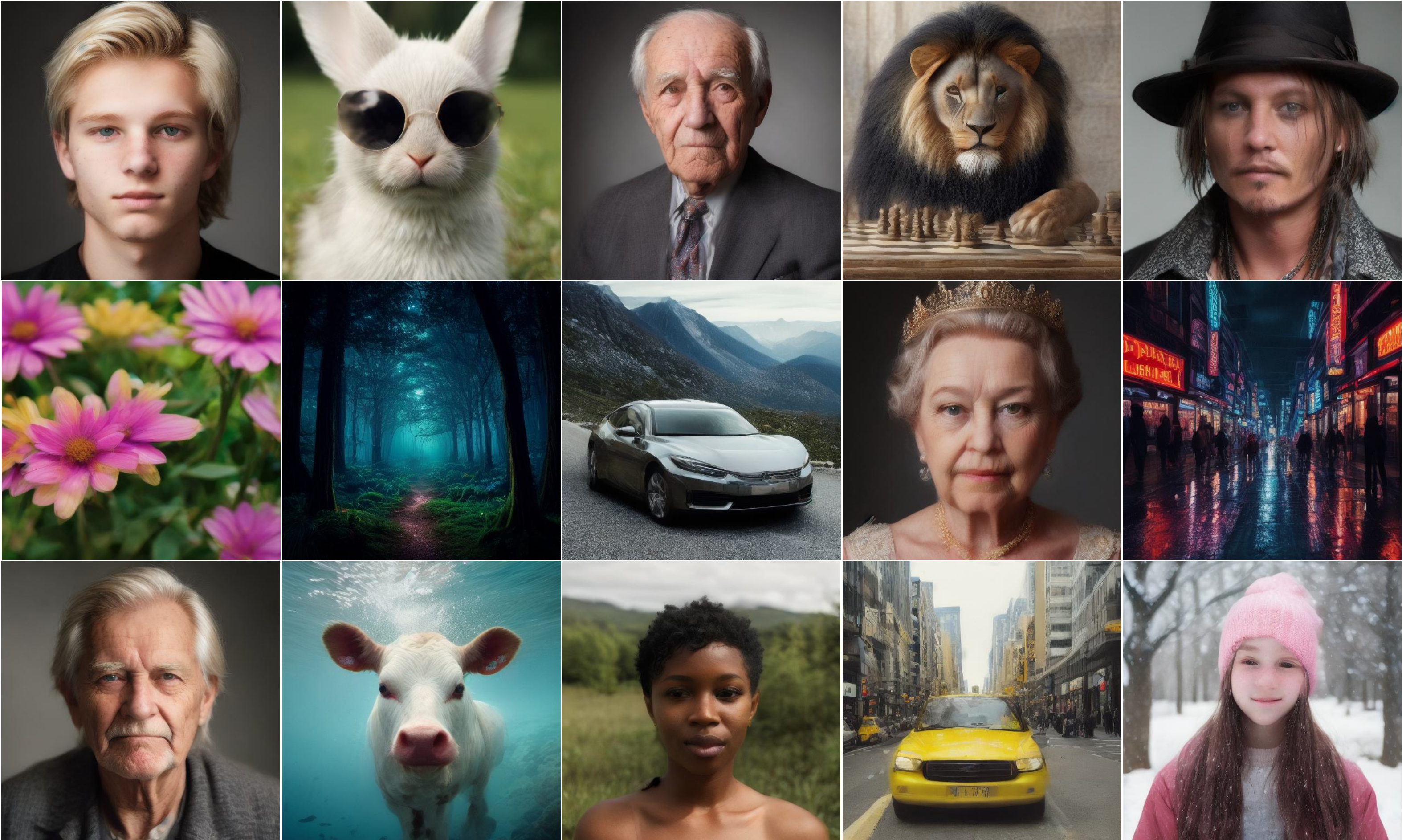}
       \vspace{-0.2in}
    \caption{\textbf{Output examples from \ours with 2 step inference.} Prompts (left to right, top to bottom): \textit{``photo portrait of a young man with blond hair''}, \textit{``a cute rabbit wearing sunglasses, photo-realistic''}, \textit{``photo portrait of an old man wearing a suit''}, \textit{``a lion playing chess, photo-realistic''}, \textit{``photo portrait of Johnny Depp, perfect skin, hyper detailed, ultra high res, RAW''}, \textit{``close up on colorful flowers''}, \textit{``an enchanted forest, trees aglow with bioluminescent fungi, mythical creatures flitting through the shadows''}, \textit{``a beautiful car in a mountain''}, \textit{``beautiful photo portrait of the queen, high resolution, 4k''}, \textit{``a bustling city street at night, neon lights reflecting off wet pavement, the hum of activity never ceasing''}, \textit{``photo portrait of an old man with blond hair''}, \textit{``a cow swimming under the sea, photo-realistic''}, \textit{``photo portrait of a black woman with short hair, nature in the background''}, \textit{``A yellow taxi driving through the city streets.''}, \textit{``photo portrait of a teenager with a pink beanie, a pop of color on a snowy day''}.}\label{fig:supp_edgefusion_2step}
\end{figure*}

%% file: figure/supp_edgefusion_4step.tex
\begin{figure*}[t]
  \centering
  \includegraphics[width=\linewidth]{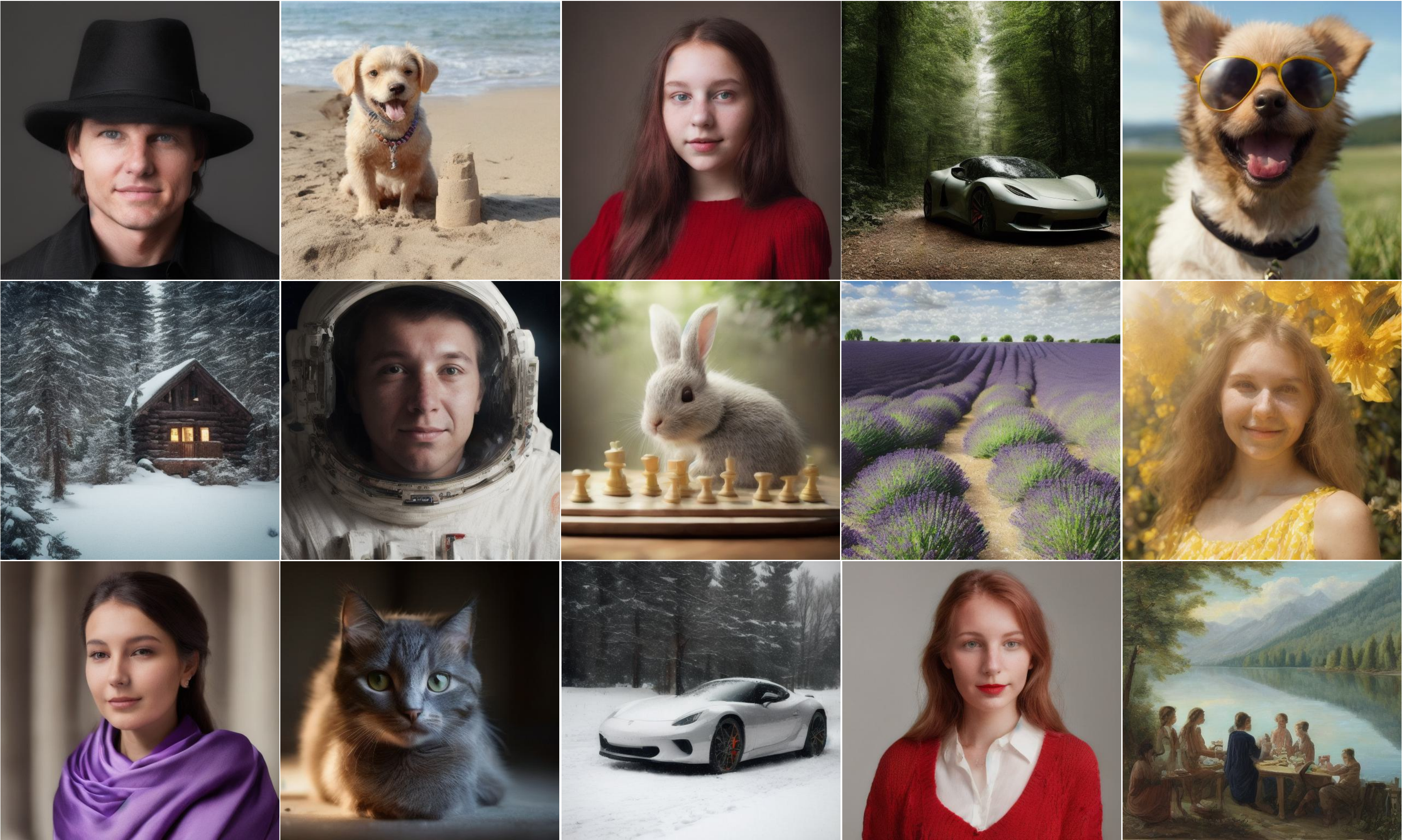}
       \vspace{-0.2in}
    \caption{\textbf{Output examples from \ours with 4 step inference.} Prompts (left to right, top to bottom): \textit{``photo portrait of Tom Cruise wearing a black hat''}, \textit{``a cute dog making a sandcastle, photo-realistic''}, \textit{``photo portrait of a young woman dressed in red''}, \textit{``a beautiful sports car in a forest''}, \textit{``a cute dog wearing sunglasses, photo-realistic''}, \textit{``a cabin in snowy forest''}, \textit{``photo portrait of an astronaut, high resolution, 4k''}, \textit{``a cute rabbit playing chess, photo-realistic''}, \textit{``a field of lavender swaying, the scent perfuming the air, bees buzzing lazily from flower to flower''}, \textit{``photo portrait of a woman in a golden yellow summer dress, capturing the essence of sunny days''}, \textit{``a woman in a purple silk scarf, adding a touch of elegance to her outfit, high resolution, 4k''}, \textit{``a cute cat, hyper detailed, cinematic light, ultra high res, best shadow, RAW''}, \textit{``a beautiful sports car under the snow''}, \textit{``photo portrait of a woman in a bright red cardigan, buttoned up over a white blouse''}, \textit{``painting of a group of people eating next to a lake''}.}\label{fig:supp_edgefusion_4step}
\end{figure*}